%% file: main.tex
\documentclass[letterpaper, 10 pt, conference]{ieeeconf}  
\IEEEoverridecommandlockouts                              
\usepackage{packages}

\title{\LARGE\bf 
From Screen to Stage: Kid Cosmo, A Life-Like, Torque-Controlled Humanoid for Entertainment Robotics

}
\author{
    Havel Liu$^{1*}$, Mingzhang Zhu$^{1*}$, Arturo Moises Flores Alvarez$^{1}$, Yuan Hung Lo$^{1}$, \\
    Conrad Ku$^{1}$, Federico Parres$^{1}$, Justin Quan$^{1}$, Colin Togashi$^{1}$, \\
    Aditya Navghare$^{1}$, Quanyou Wang$^{1}$, and Dennis W.~Hong$^{1}$%
    \thanks{$^{1}$Department of Mechanical and Aerospace Engineering, University of California, Los Angeles (UCLA), Los Angeles, CA, USA.}%
    \thanks{Emails: {\tt\small havel@g.ucla.edu, normanzmz@g.ucla.edu}}%
    \thanks{*Equal contribution.}%
}

\begin{document}
\maketitle
\thispagestyle{empty}
\pagestyle{empty}

\begin{abstract}
Humanoid robots represent the cutting edge of robotics research, yet their potential in entertainment remains largely unexplored. Entertainment as a field prioritizes visuals and form, a principle that contrasts with the purely functional designs of most contemporary humanoid robots. Designing entertainment humanoid robots capable of fluid movement presents a number of unique challenges. In this paper, we present Kid Cosmo, a research platform designed for robust locomotion and life-like motion generation while imitating the look and mannerisms of its namesake character from Netflix's \textit{The Electric State}. Kid Cosmo is a child-sized humanoid robot, standing 1.45 m tall and weighing 25 kg. It contains 28 degrees of freedom and primarily uses proprioceptive actuators, enabling torque-control walking and lifelike motion generation. Following worldwide showcases as part of the movie's press tour, we present the system architecture, challenges of a functional entertainment robot and unique solutions, and our initial findings on stability during simultaneous upper and lower body movement. We demonstrate the viability of performance-oriented humanoid robots that prioritize both character embodiment and technical functionality.

\end{abstract}


\input{Sections/S1_introduction}
\input{Sections/S2_design_of_COSMO}
\input{Sections/S3_software_and_control}

\input{Sections/S4_results}

\input{Sections/S5_conclusion}

{
\bibliographystyle{IEEEtran}
\bibliography{reference_updated}
}

\end{document}

%% file: Sections/S1_introduction.tex
\section{Introduction}
\label{sec:intro}

Robotics is gaining an increasing amount of interest as a way to enhance productivity, improve quality of life, and even provide entertainment. However, a clear distinction exists between robotics for entertainment and the cutting edge of legged robotics. Robotics for entertainment tend to focus mainly on how the robot looks and performs certain entertainment tasks \cite{lau_integrating_2025}, while humanoid robots generally prioritize functional aspects such as a mass-optimized design and robust locomotion \cite{kojima_robot_2019}.

Robotics in entertainment started mainly as animatronics, which are mechanized puppets designed to look and move like a particular person or character. Early work focused on generating movement \cite{hollerbach_anthropomorphic_1996} and reaching a level of realism by crossing the uncanny valley \cite{oh_design_2006}. These early robots operated in very controlled environments and possessed very limited mobility. As development in entertainment robotics pushed toward more functionality, we saw the rise of robots capable of playing musical instruments \cite{solis_development_2009}, playing games with humans \cite{johnson_exploring_2016}, and expressive motion \cite{grandia_design_2024}.

In contrast, early humanoid robot research focused on platforms capable of stable movement such as ASIMO \cite{sakagami_intelligent_2002} and HRP-2 \cite{kaneko_humanoid_2004}. These robots used actuators with high transmission ratios to achieve the necessary joint torques for locomotion. Although this results in reasonable dynamic properties for smaller humanoids such as DarwIn-OP \cite{ha_development_2011} and NAO \cite{gouaillier_mechatronic_2009}, the high reflected inertia and slower accelerations of these actuators hamper the ability of larger robots to perform highly dynamic movements and reject environmental disturbances. In response, the design of legged robot actuators trended toward lower transmission ratio and higher torque actuators known as proprioceptive actuators \cite{wensing_proprioceptive_2017}. These actuators allow for higher control bandwidth \cite{katz_mini_2019} and have been successfully implemented in robots such as BRUCE \cite{liu_design_2022} and the MIT Humanoid \cite{saloutos_design_2023}. The use of proprioceptive actuators has also enabled torque-controlled locomotion \cite{ahn_development_2023}, further enhancing dynamic capabilities.

\begin{figure}[t!]
    \centering
    \begin{minipage}[b]{0.48\linewidth}
        \centering
        \includegraphics[width=\linewidth]{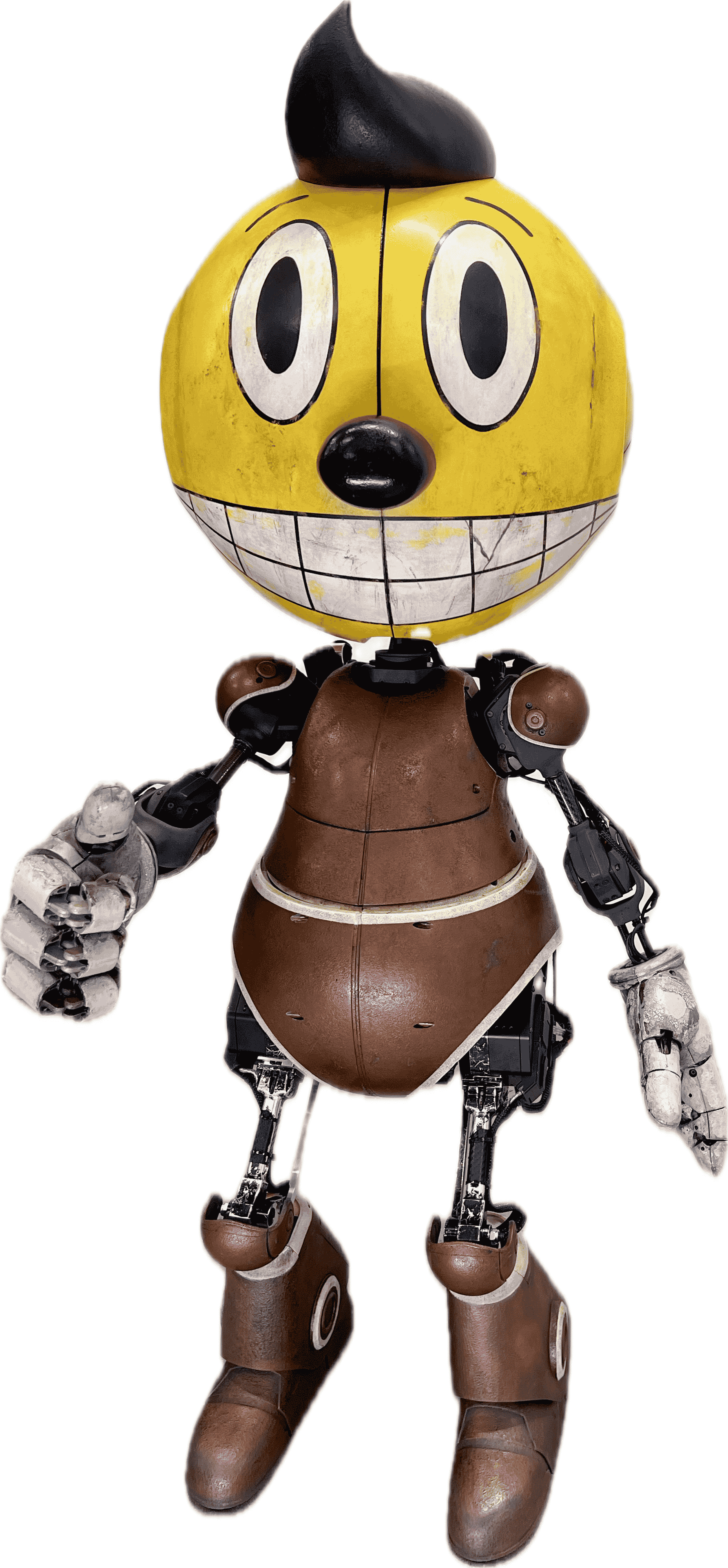}
        \caption*{(a)}
    \end{minipage}
    \hspace{1em}
    \begin{minipage}[b]{0.29\linewidth}
        \centering
        \includegraphics[width=\linewidth]{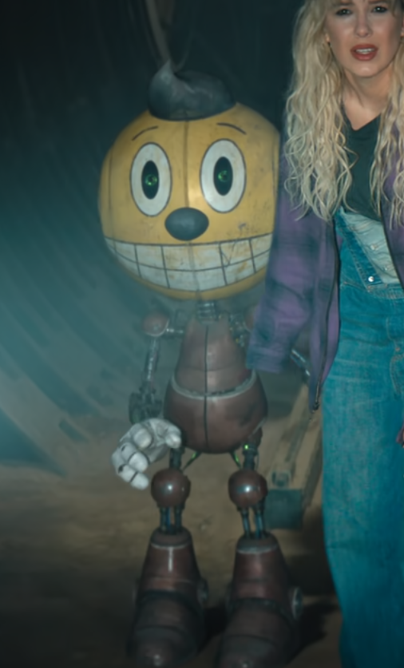}
        \caption*{(b)}
        \vspace{0.5em}
        \includegraphics[width=\linewidth]{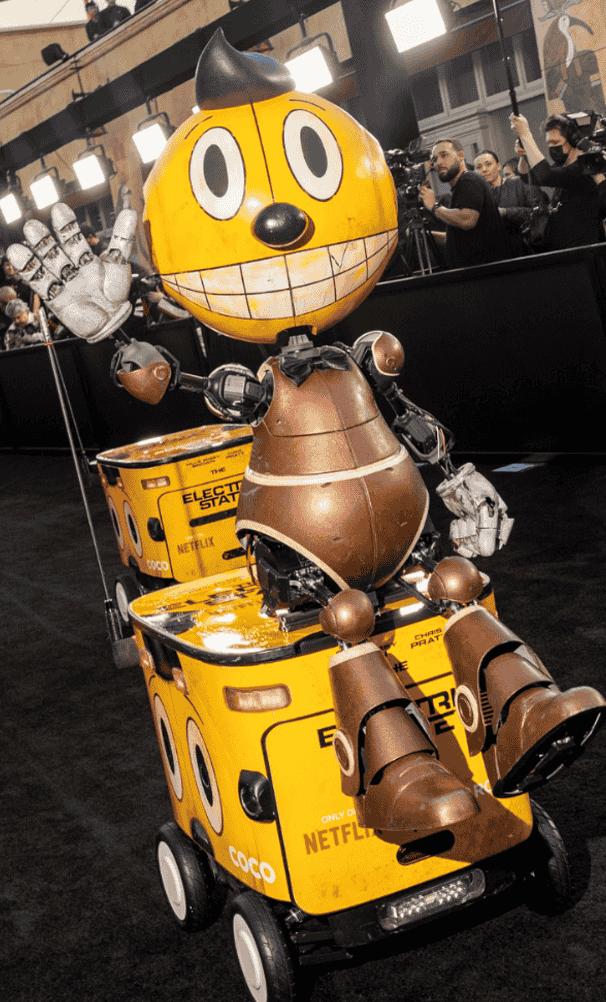}
        \caption*{(c)}
    \end{minipage}
    \caption{The assembled robot Kid Cosmo (a), the character from \textit{The Electric State} (b), and Kid Cosmo sitting and performing upper body motion at the movie premiere (c).}
    \label{fig:action_shot}
\end{figure}

Entertainment robotics can benefit from the recent advancements in humanoid robot design to create a new generation of robots capable of immersive human interaction and real world performance. In this paper we present Kid Cosmo (Cosmo), a child-sized humanoid robot based on its namesake character from Netflix's \textit{The Electric State} that embodies its namesake character through look and fluid, dynamic movement. Our contributions are as follows:

\begin{itemize}
    \item The use of compliance on aesthetic components to mitigate impact during dynamic movements.
    \item The unique configuration of internal components to conform to the aesthetics of the character.
    \item Validation of dynamic locomotion with disturbance rejection and simultaneous upper body gestures.
\end{itemize}


%% file: Sections/S2_design_of_COSMO.tex
\section{Design of Cosmo}
\label{sec:design of COSMO}

\begin{figure}[t!]
 \centering
 \includegraphics[width=0.8\linewidth]{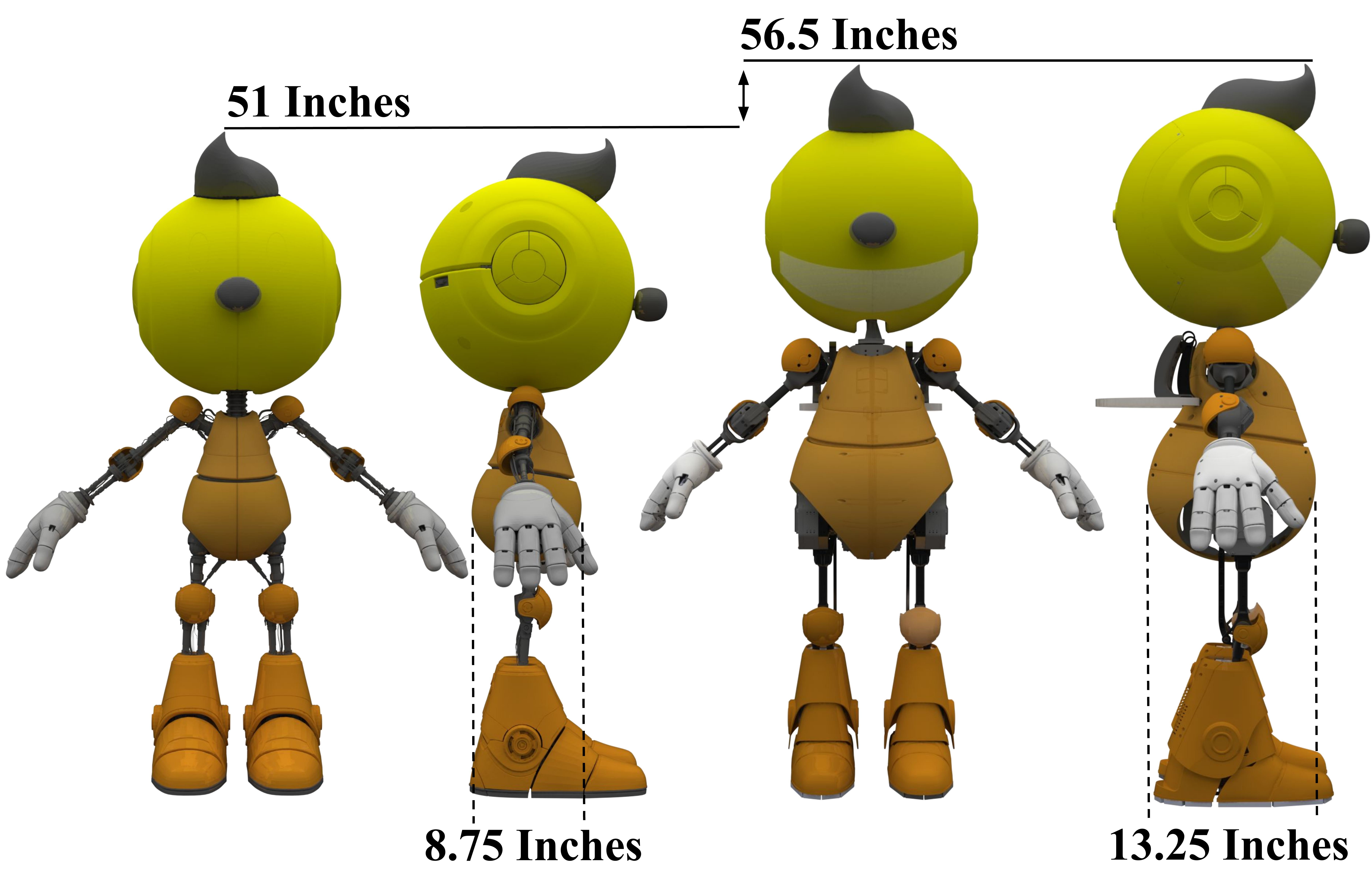}
 \caption{Size comparison between the character (left) and Cosmo (right). The robot's torso is scaled up in all axes from the character's by 150\%. All other dimensions remain identical.}
 \label{fig:comparison_figure}
\end{figure}

\begin{table}[t!]
 \centering
 \caption{Shell Mass Distribution}
 \begin{tabular}{ c c c c c c }
    & \textbf{Extremities (Head)} & \textbf{Body} & \textbf{Boots} & \textbf{Hands} \\
    \specialrule{.15em}{.05em}{0em}
    Mass [kg] & 4.40 (4.0) & 1.48 & 1.50 & 0.83
 \label{table:shell_weights}
 \end{tabular}
\end{table}

The fully assembled robot, its movie counterpart, and the robot at the movie premiere are shown in \cref{fig:action_shot}. The primary design objective of Cosmo is to provide entertainment through life-like mimicry of the character's movements and mannerisms. To achieve this goal, we focus on three areas of interest: shell, joint, and electronics design.

\subsection{Shell Design}
Given that the shells are the most prominent features of the character, keeping the shell design as similar as possible became the dominant constraint for the rest of the mechanical system. We received a visual effects (VFX) model at the beginning of the design process which served as our reference.  We split the character's shells into four categories: extremities (knees, elbows, shoulders, and head), body, boots, and hands. The shells were fabricated by W\={e}t\={a} Workshop, and their mass distribution is shown in \cref{table:shell_weights}.

The extremity shells were the easiest to design, as their original placement posed few issues. The knee, elbow, and shoulder shells do limit range of motion, but not enough to severely impact the range of gestures the robot can perform. The character's large head raised concerns about mass and fragility during locomotion, so the head is made of thin 3D printed nylon and reinforced with a carbon fiber inner structure to resist minor impact.

\begin{figure}[t!]
    \centering
    \hspace{-1.5cm}
    \begin{subfigure}{0.375\linewidth}
        \includegraphics[width=1\linewidth]{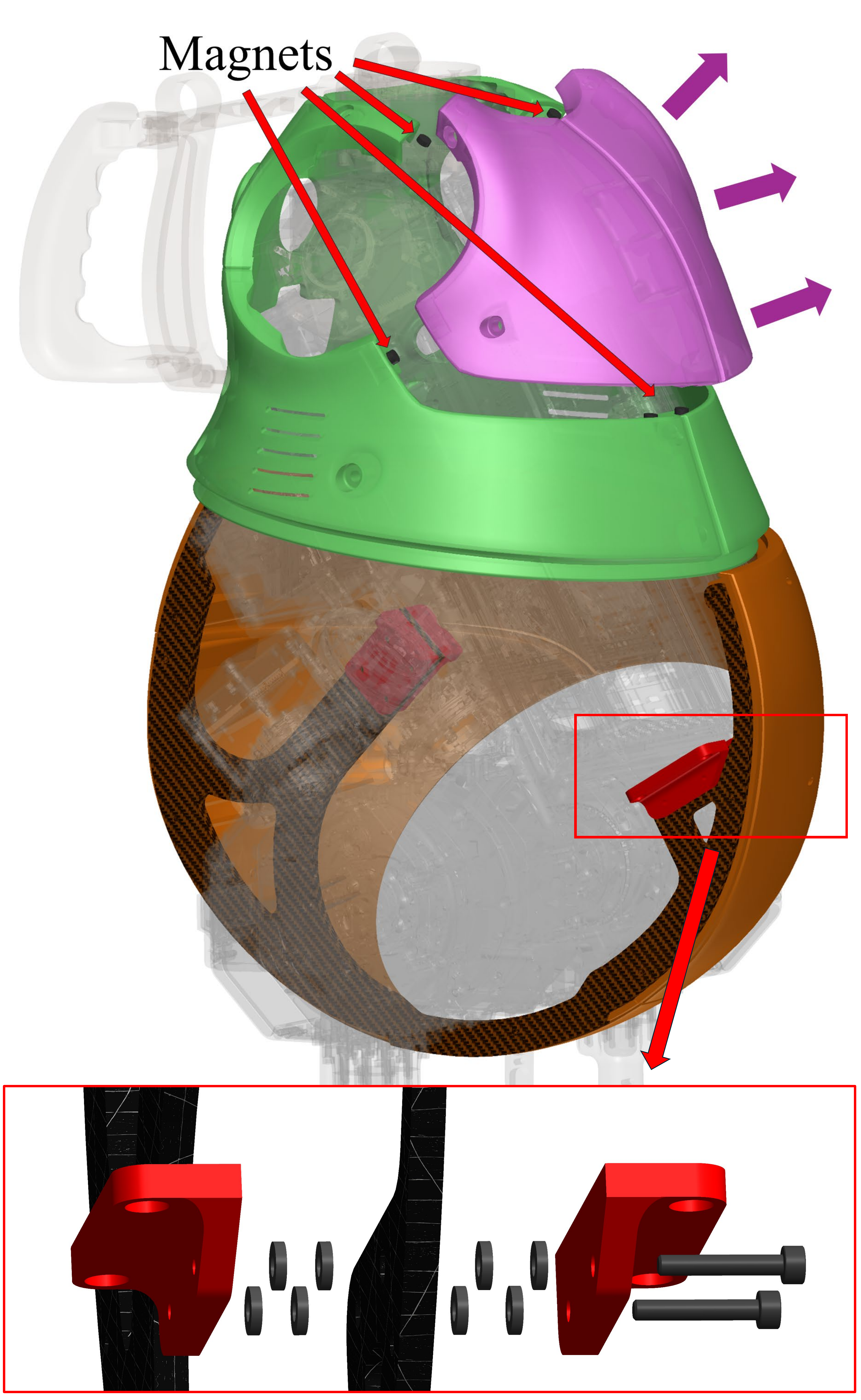}
        \caption{Body Shells}
        \label{body_shell_subfig}
    \end{subfigure}
    \begin{subfigure}{0.625\linewidth}
        \includegraphics[width=1\linewidth]{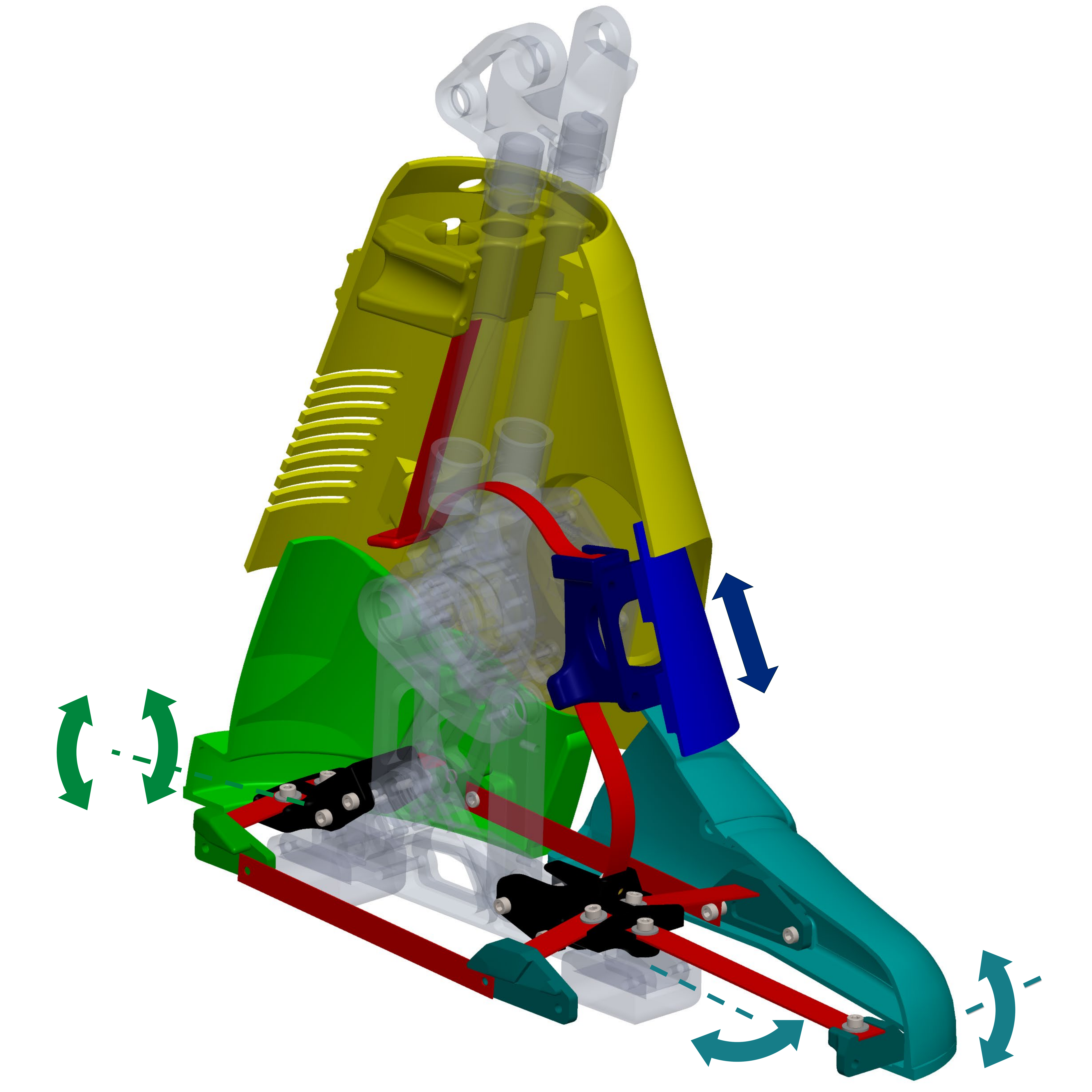}
        \caption{Boots}
        \label{boot_subfig}
    \end{subfigure}
    \hspace{-2cm}
    \label{cover_fig}
    \caption{Functional design features of the shells. The front plate (purple) detaches from the shirt (green) using magnets, allowing for quick battery access. The pants (orange) are connected to a carbon fiber plate which is attached to the torso using brackets (red). Rubber washers are sandwiched between the carbon fiber and the brackets, introducing compliance. The boot shells are connected by steel flexures (red) to allow movement in the axes shown.}
\end{figure}

The torso contains most of Cosmo's required electronic components, so volume was the biggest issue with the body shell design. The torso of the VFX model was not large enough to fit the electronics and was scaled up to 150\% to allow valid packaging. The torso is the only dimensional change, so Cosmo and the character remain proportionally similar. A comparison of the two is shown in \cref{fig:comparison_figure}. The VFX model also contains a small bottom hole for each of the spherical hip joints which is insufficient for the robot. Thus, the bottom openings on the robot's shells are enlarged to not only fit the hip and knee actuators but also to allow for sufficient workspace for locomotion. Cosmo is designed for use in live entertainment settings where prolonged use may be required, so quick servicing is necessary. Thus, the front of the body shells use a set of magnets to quickly detach from the torso to allow for battery access. For dynamic movement, our main concern was the impact between the legs and the body shells during locomotion, so rubber washers were used to introduce compliance between the torso and the body shells. These design details are shown in \cref{body_shell_subfig}.

The boot design focused on creating a realistic boot without restricting motion during locomotion. Due to the large profile of the boots, any footstep not parallel to the contact surface could cause damage and/or instability, resulting in the need for a compliant design. Spring steel flexures were used to not only position the different shells relative to the foot but to facilitate relative movement between the shells of the boot and the foot, reducing the likelihood of damage. The components of the boot and their movement axes are pictured in \cref{boot_subfig}. During ankle rotation, the dark blue plate slides up and down to cover the gap between the yellow and teal piece, resulting in a fully formed boot with full range of motion. During locomotion, the green and teal piece can deflect to prevent damage.

Our goal with the hand design was to combine fully mobile fingers with the character's distinctive thick and silver look. To shorten development time, we purchased a prosthetic hand from the company Mand.ro and replaced the covers with our own shells. The shell of each finger digit can slide around each other, allowing the fingers to fully open and close. Unlike the rest of the shells, the hand shells were manufactured and acrylic painted in-house.

\begin{figure}[t!]
 \centering
 \includegraphics[width=1\linewidth]{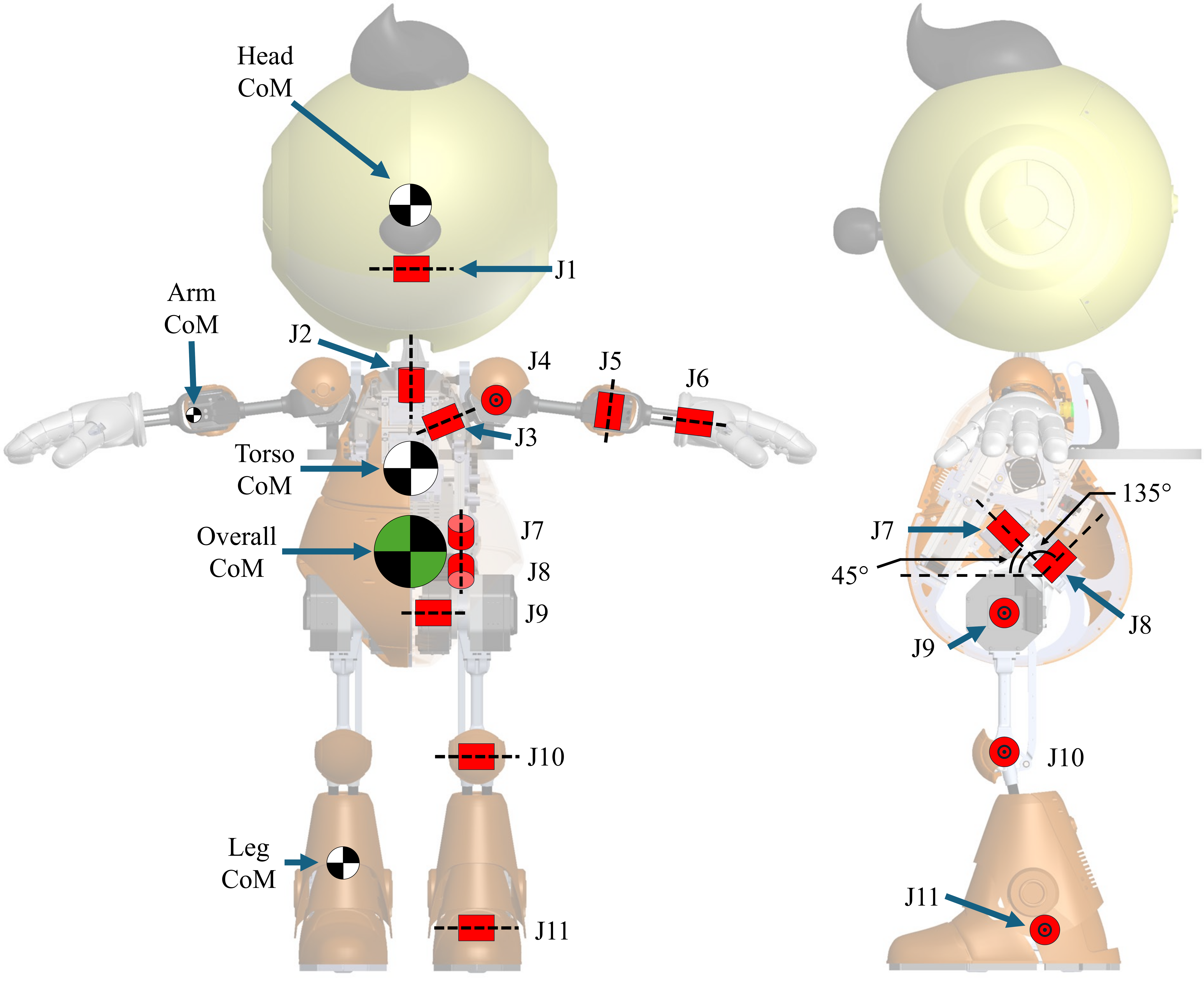}
 \caption{Complete joint configuration of Cosmo detailing the location and orientation of the joints in the head, arms, and legs. The hand contains 4 degrees of freedom that are not pictured for clarity. The link and overall center of masses are also shown, with the size of the symbol indicating relative size.}
 \label{fig:configuration_figure}
\end{figure}

\begin{table}[t!]
 \centering
 \caption{Joint and Link Specifications}
 \begin{tabular}{ l c c c c c }
    \textbf{Link} & \textbf{Mass [kg]} & \textbf{Joint} & \textbf{Name} & \textbf{Actuator} & \textbf{Limits [deg]} \\
    \specialrule{.15em}{.05em}{0em}
    \multirow{2}{*}{Head} & \multirow{2}{*}{4} & J1 & Pitch & KBMB & [-30 30] \\
    & & J2 & Yaw & KBMB & [-90 90] \\ \hline
    \multirow{4}{*}{Arm} & \multirow{4}{*}{1.5} & J3 & Pitch & KBMB & [-190 90] \\
    & & J4 & Roll & KBMB & [-85 35] \\
    & & J5 & Elbow & KBMB & [-25 115] \\
    & & J6 & Wrist & Micro & [-180 180] \\ \hline
    \multirow{5}{*}{Leg} & \multirow{5}{*}{2.5} & J7 & Hip 1 & KBMB & [-40 20] \\
    & & J8 & Hip 2 & KBMB & [-45 40] \\
    & & J9 & Pitch & Panda Plus & [-130 55] \\
    & & J10 & Knee & Panda Plus & [-90 0] \\
    & & J11 & Ankle & KBMB & [-25 40] \\ \hline
    Torso & 13 & - & - & - & - 
 \label{table:joint_table}
 \end{tabular}
\end{table}

\subsection{Joint Design}
Cosmo's unique joint configuration is driven by the character's design as well as the volume constraints of the torso. The joint configuration of the robot and center of masses of the links can be seen in \cref{fig:configuration_figure} while its joint details and link masses are shown in \cref{table:joint_table}. For clarity, the finger joints are omitted and only the left side joints are shown. Cosmo’s legs have only 5 degrees of freedom, intentionally omitting the ankle roll joint. This design choice reduces the distal mass of the feet, which helps the leg accelerate faster during the swing phase and better matches the simplified model used in the locomotion control.

To minimize hip volume, a unique configuration is used for the hip yaw and hip roll joints. Instead of the usual 0$\degree$ and 90$\degree$ configuration relative to the X-axis \cite{ficht_bipedal_2021}, a 45$\degree$ and 135$\degree$ configuration is used to couple the two axes and their loads together, allowing identical, smaller actuators to be used for both. This angled joint arrangement also packages well in the contours of the body shells. The hip joint configuration angles are shown in \cref{fig:configuration_figure}.

Due to the need for high control bandwidth and smooth motion we elected to use proprioceptive actuators, with the Panda Bear Plus (Panda Plus) and Koala Bear Muscle Build (KBMB) from Westwood Robotics being chosen. The Panda Plus is used for the high load hip pitch and knee joints, while the KBMB is used for hip 1, hip 2, and ankle joints since they have similarly lower loads. Notably, the knee joint loads are too high for a Panda Plus to handle due to thermal constraints, so a 3:1 planetary gearbox was added. The tradeoff in slightly lower efficiency and larger reflected inertia is worth the lower average motor torque. The upper body also uses KBMBs to achieve smooth motion and minimize system complexity. Due to packaging constraints, the wrist joint and fingers use brushed micro gearmotors from Pololu. Details of these actuators can be seen in \cref{table:actuator_table}.

\begin{figure}[t!]
 \centering
 \includegraphics[width=1\linewidth]{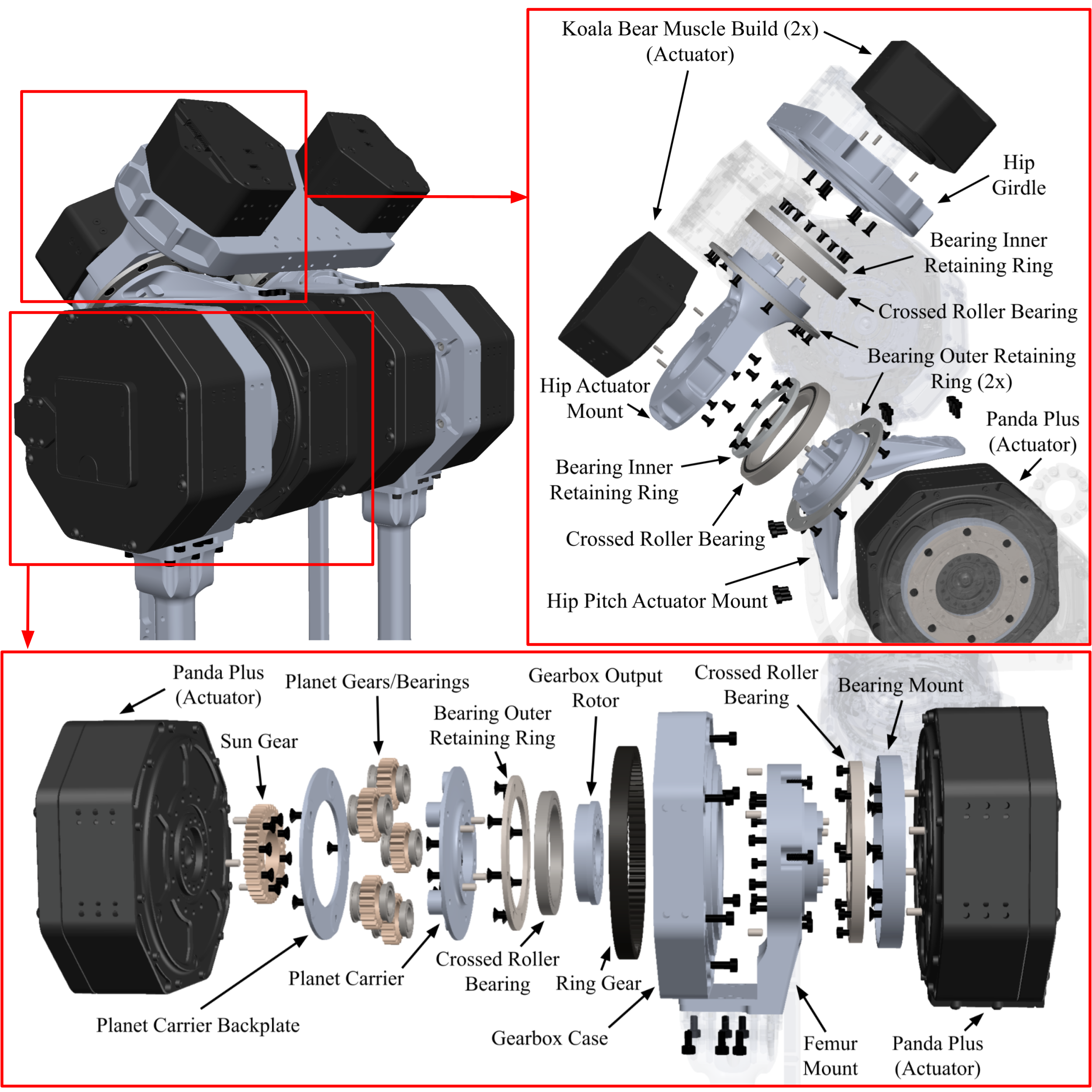}
 \caption{Explosion views of the hip and knee joints. Depicted is the actuator torque transfer and bearing constraint strategy as well as the custom planetary transmission of the knee.}
 \label{fig:explosion_figure}
\end{figure}

In order to minimize limb inertia and maximize dynamic performance, the leg actuators are placed as close to the torso as possible. Thus, the knee actuator is placed coaxially with the hip pitch actuator and transfers torque to the knee joint using a parallel four-bar linkage. To minimize hip width and match the character's appearance, crossed-roller bearings are used for the hip and knee degrees of freedom for minimal thickness load and moment handling. An explosion of the hip and knee joint assemblies are shown in \cref{fig:explosion_figure}.

\begin{table}[t!]
 \centering
 \caption{Actuator Specifications}
 \begin{tabularx}{\linewidth}{ l c c c }
 \textbf{Specification} & \multicolumn{1}{c}{\textbf{KBMB}} & \multicolumn{1}{c}{\textbf{PB Plus}} & \multicolumn{1}{c}{\textbf{Micro Gearmotor}} \\
 \specialrule{.15em}{.05em}{0em}
 Mass (g) & 285 & 925 & 9.5 \\
 \hline
 Gear Ratio & 20:1 & 10:1 & 298:1 \\
 \hline
 Stall Torque (N$\cdot$m) & 8 & 26.5 & 0.32 \\
 \hline
 Torque Constant (N$\cdot$m/A) & 1.16 & 1.3 & 0.43 \\
 \hline
 Speed Constant (RPM/V) & 9 & 7.1 & 9.2 \\
 \hline
 \label{table:actuator_table}
 \end{tabularx}
\end{table}

\subsection{Electronics Design}
Cosmo's electronics system includes two 22.2 V batteries, a mini PC and associated voltage converter for computation, a 3DM-CV7-AHRS Inertial Measurement Unit (IMU) for state estimation, and a wired and wireless emergency stop (Estop). The batteries are placed behind the removable front cover, allowing for easy access during events. Foot contact sensors are placed in the heel and toe of each foot, allowing for discrete detection of contact during locomotion. The wrist joint and fingers are controlled using a wrist control board that handles power conversion, communication, and closed-loop control. The contact sensors and actuators communicate with the mini PC using an RS-485 bus.

Since the robot is often operated at live events with hundreds of people, it commonly experiences heavy wireless interference and, at some venues, active jamming in the 2.4 GHz frequency band. This eliminated common Bluetooth controllers as a means of wireless control, so we developed a custom 915 MHz wireless controller for use during events. The controller includes 10 buttons, 2 analog joysticks, a screen, and a wireless transmitter. The lower frequency of the controller also greatly increased the operating range, an important benefit in some events where the visibility of event staff, including robot operators, is undesirable. The receiver for the controller and wireless Estop are placed within the body behind the batteries.

\begin{figure}[t!]
 \centering
 \includegraphics[width=1\linewidth]{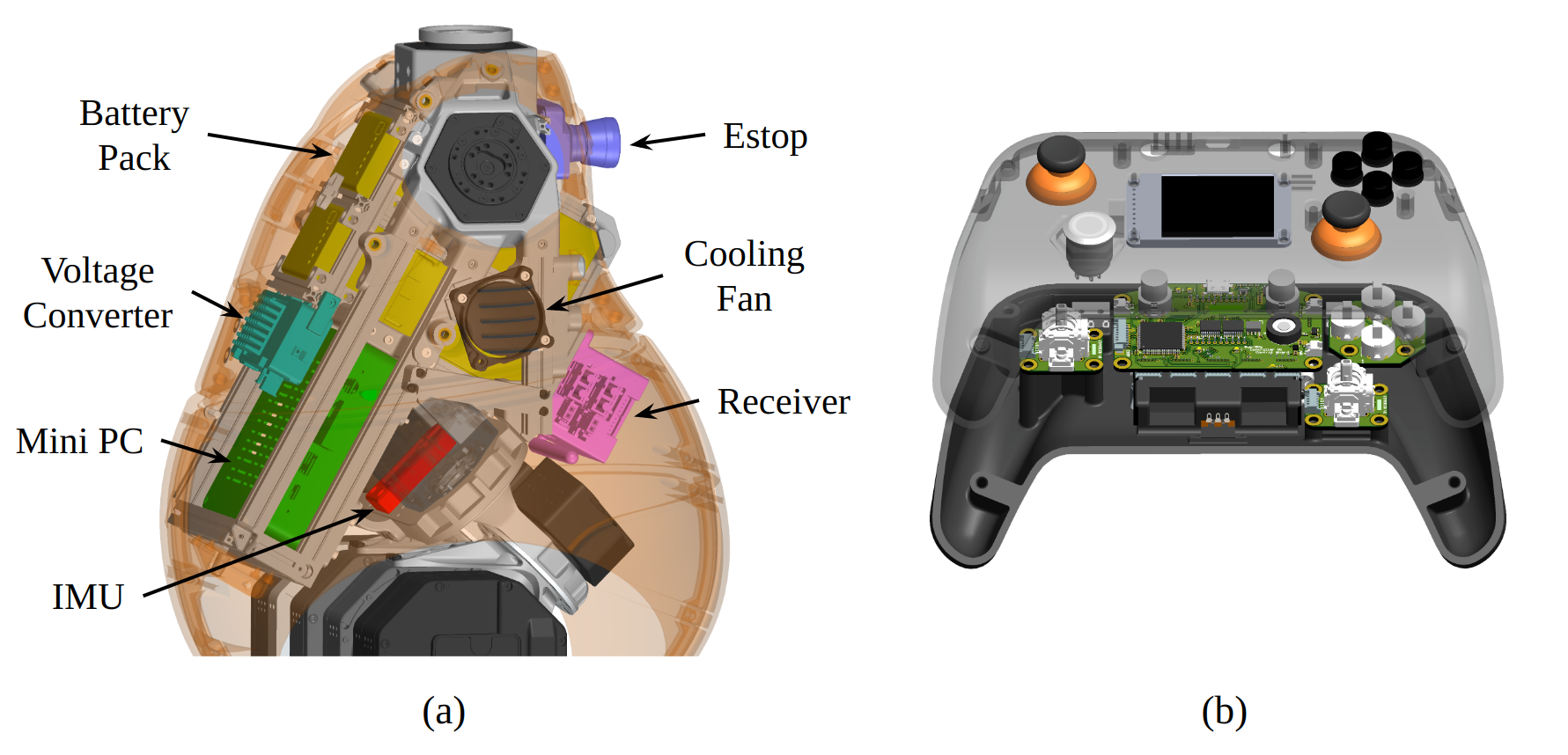}
 \caption{Layout of Cosmo's electronics system within the torso (left) and inside view of the custom wireless controller (right).}
 \label{fig:electronics_figure}
\end{figure}

All electronics mentioned in this section besides the mini PC and IMU were created by our lab. The electronics system layout and controller details can be seen in \cref{fig:electronics_figure}.

%% file: Sections/S3_software_and_control.tex
\section{Software and Control}
\label{sec:software and control}
\begin{figure*}[t!]
    \centering
    \includegraphics[width=0.9\linewidth]{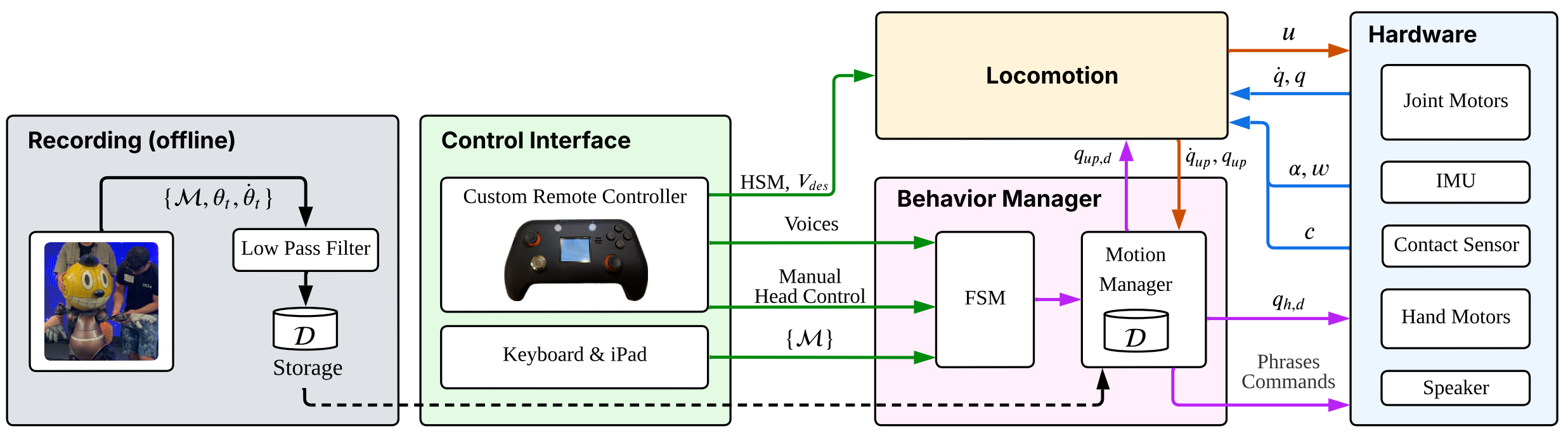}
    \caption{The software architecture of Cosmo.}
    \label{fig:software_pipeline}
\end{figure*}

The high-level software architecture of Cosmo is shown in Fig.~\ref{fig:software_pipeline}. The first phase, pre-recording, involves human operators manually manipulating the robot’s torque-disabled limbs to create a set of desired motions, $\mathcal{D}$. A custom remote controller and an Xbox controller are used to control manual head movements, trigger voice outputs, and command the desired robot velocity, $V_{cmd}$. A keyboard and iPad are used to select specific motions from $\mathcal{D}$. The behavior manager processes these inputs, sending phrase commands and hand motor position commands, $q_{h,d}$, directly to the speaker and hand motors. Commanded upper body joint positions, $q_{up,d}$, are sent to the shared memory within the locomotion stack as reference trajectories. The locomotion stack then processes joint positions $\mathbf{q}$, joint velocities $\mathbf{\dot{q}}$, contact sensor readings $\mathbf{c}$, and IMU measurements $\boldsymbol{\alpha}$ and $\boldsymbol{\omega}$ to estimate the robot's state. Finally, the desired torque commands $\mathbf{u}$ are generated and sent to the joint motors for execution.

\begin{figure}[htbp]
    \centering
    \includegraphics[width=0.9\linewidth]{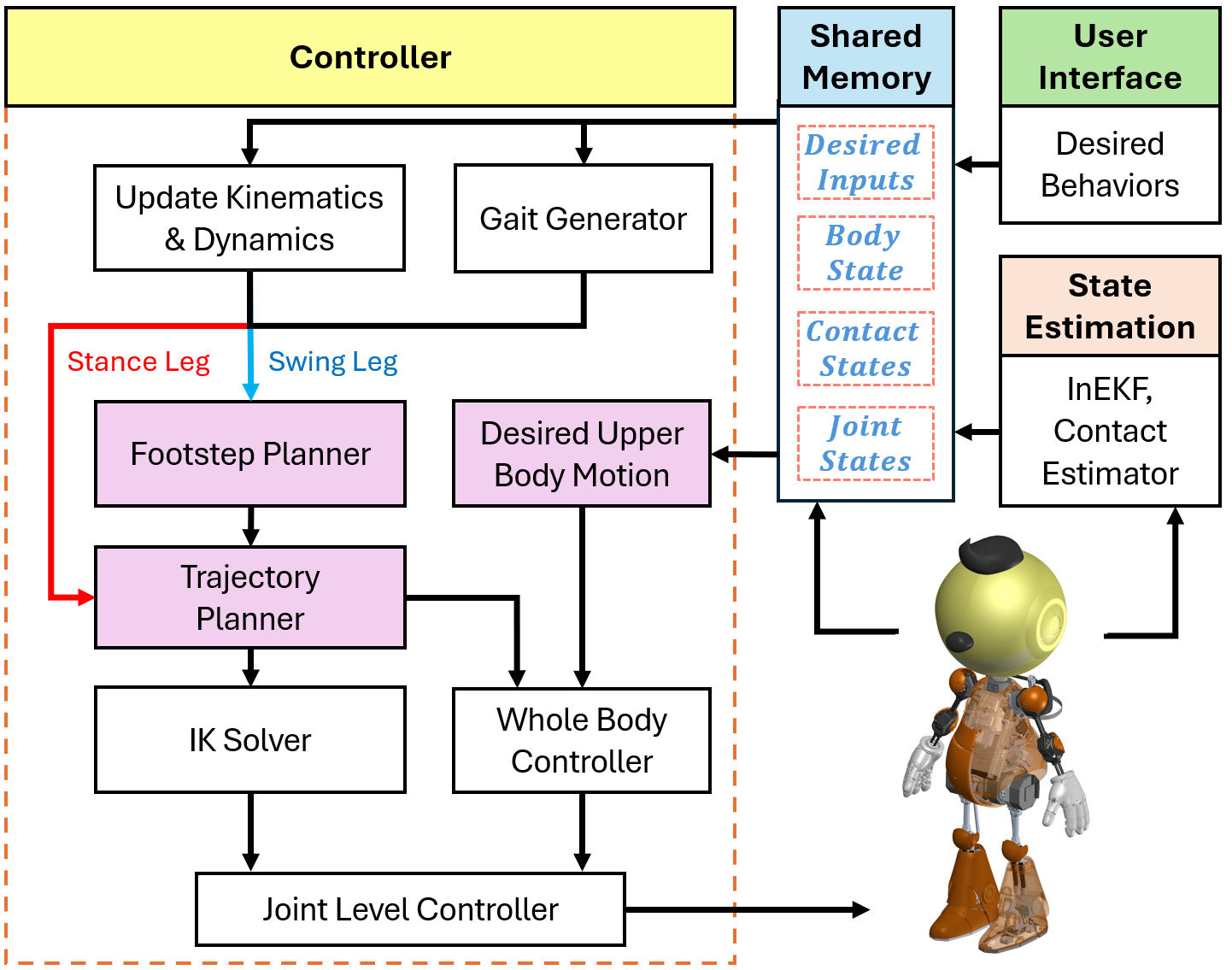}
    \caption{The locomotion framework of Cosmo. Purple blocks represent the reference trajectory planners.}
    \label{fig:locomotion_framework}
\end{figure}

\subsection{Locomotion Overview}
Cosmo’s locomotion framework (Fig.~\ref{fig:locomotion_framework}) is a hierarchical, model-based architecture that enables dynamic walking coordinated with upper body motion. High-level behavior commands from the user interface are transmitted to shared memory, which stores desired inputs, joint states, contact states, and body estimates. State estimation is performed using a Contact-Aided Invariant Extended Kalman Filter (InEKF) and a contact estimator \cite{hartley_contact-aided_2019}.

The controller updates the robot’s kinematics and dynamics based on sensor feedback and uses a symmetric, contact-based gait generator \cite{togashi_control_2024} to determine each leg’s stance or swing phase according to the desired gait period (0.3–0.5s). For swing legs, a footstep planner computes target touchdown position, and a trajectory planner generates smooth swing foot motions. During stance, foot trajectories maintain zero acceleration. Desired upper body motions are also retrieved from shared memory.

All reference trajectories—including swing foot, CoM, and upper body motion—are sent to an inverse kinematics (IK) solver and a whole-body controller (WBC), which compute the desired joint positions, velocities, and torques. These joint commands are then passed through a PD controller with feedforward torque to produce the final control effort applied to the robot. 

\subsection{Reference Trajectory Generation}
\subsubsection{Linear Inverted Pendulum Model}
\begin{figure}[htbp]
    \centering
    \includegraphics[width=0.4\linewidth]{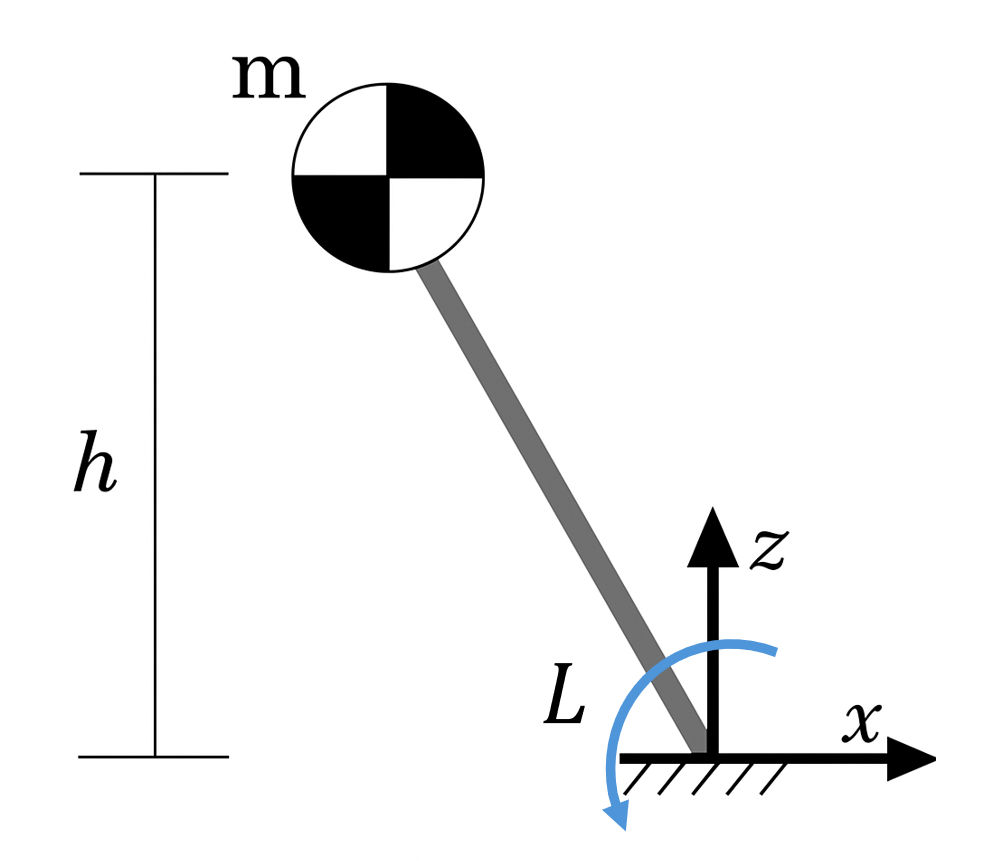}
    \caption{Linear Inverted Pendulum Model}
    \label{fig:LIP_model}
\end{figure}

To compute footstep locations, the planner uses the Linear Inverted Pendulum (LIP) model \cite{kajita_3d_2001}, shown in Fig.~\ref{fig:LIP_model}. The model assumes a point mass with massless legs and no foot slip. Cosmo’s high CoM and five-DoF legs without ankle roll align well with these assumptions. With constant height $h$, the system dynamics can be simplified to the following equation of motion:
\begin{equation}
\begin{bmatrix}
\dot{x} \\ 
\ddot{x}
\end{bmatrix}
=
\begin{bmatrix}
0 & 1 \\ 
\frac{g}{h} & 0
\end{bmatrix}
\begin{bmatrix}
x \\ 
\dot{x}
\end{bmatrix}
\end{equation}

\subsubsection{Footstep Planner}
Compared to tracking linear momentum, it has been shown that tracking the angular momentum about the contact point, $L$, provides improved stability for the LIP model, where $L = mh\dot{x}$. Following the approach in \cite{gong_one-step_2021}, the desired footstep touchdown position is computed as:
\begin{equation} p_{SF}(t_c) = \frac{L_{\text{des}}(T_{s}) - \cosh(\omega(T_{s} - t_c))L(t_c)}{mh\omega \sinh(\omega(T_{s} - t_c))}, \end{equation}

where $t_c$ is the current time, $T_s$ is the swing period, $L(t_c)$ is the current angular momentum, and $L_{\text{des}}(T_s)$ is the desired angular momentum. The natural frequency is defined as $\omega = \sqrt{\frac{g}{h}}$. The desired angular momentum can be decomposed along the $x$ and $y$ directions as:
\begin{equation} L_{x,\text{des}} = mhv_{x,\text{des}}, \end{equation}
\begin{equation} L_{y,\text{des}} = mhv_{y,\text{des}} \pm mhW \frac{\omega \sinh(\omega T_{s})}{1 + \cosh(\omega T_{s})}, \end{equation}

where $v_{\text{des}}$ denotes the desired velocity, and $W$ represents the nominal step width.

\subsubsection{Swing Leg Trajectory}
After determining the next touchdown position, a smooth trajectory is required to guide the foot to the target location. To breakdown frictional constraints, the foot must rise above a specified step height during the swing phase of $Z$ direction trajectory. To satisfy both endpoint and midpoint constraints, a sixth-order polynomial trajectory is generated as follows:
\setlength{\abovedisplayskip}{4pt}
\begin{equation}
\small
p(\phi) = \sum_{i=0}^{6} c_i \phi^i
\end{equation}
\setlength{\belowdisplayskip}{4pt}

where phase parameter $\phi \in [0,1)$. The coefficients are computed to satisfy boundary conditions on position, velocity, and acceleration at both the start and end of the swing phase (i.e., zero initial and final velocity and acceleration). Additionally, a midpoint constraint is imposed to enforce a peak step height: $p(0.5) = p_{h,\text{step}}$.

\subsubsection{Body Trajectory}
The body reference trajectory includes the desired CoM velocity, body yaw rate, and body orientation, primarily determined by the commanded velocity ($V_x$, $V_y$) and yaw rate. The desired body yaw is set midway between the left and right foot yaws to maintain balance.

\subsection{Whole Body Control}
To compute the appropriate motor torques for tracking the desired swing and stance trajectories as well as other desired tasks, WBC is widely used as a model-based approach for dynamic locomotion \cite{ahn_development_2023, ding_orientation-aware_2022}.

\subsubsection{Full-Order Model}
To accurately compute joint torques, a full-order rigid-body dynamics model is used instead of the simplified LIP model. The equation of motion is:
\setlength{\abovedisplayskip}{4pt}
\begin{equation}
\mathbf{H(q)\ddot{q} + C(q, \dot{q}) = S^{\top}\tau + J_{c}^{\top}(q)f}
\end{equation}
\setlength{\belowdisplayskip}{4pt}
where $\mathbf{q = [q_b^\top, q_j^\top]^\top}$ is the generalized coordinate vector, with $\mathbf{q_b} \in \mathbb{R}^{6 \times 1}$ representing the body pose and $\mathbf{q_j} \in \mathbb{R}^{18 \times 1}$ representing the actuated joint positions. $\mathbf{H} \in \mathbb{R}^{18 \times 18}$ is the mass matrix, and $\mathbf{C} \in \mathbb{R}^{18 \times 1}$ is the vector of Coriolis and gravitational forces. $\mathbf{S} = [\mathbf{0}^{18 \times 6}, \mathbf{1}^{18 \times 18}]$ is the actuation selection matrix. $\mathbf{J_C} \in \mathbb{R}^{4 \times 18}$ denotes the contact Jacobian matrix, and $\mathbf{f} \in \mathbb{R}^{4 \times 1}$ is the contact force vector.

\subsubsection{Tasks}
The primary goal of WBC is to track desired tasks, which, in this paper, is achieved by regulating task-space accelerations $\mathbf{\ddot{x}_{\text{des}}}$. This approach allows the controller to simultaneously ensure dynamic consistency, enforce contact constraints, and coordinate multiple tasks within a unified optimization framework. A PD controller around the desired position $\mathbf{x_{\text{des}}}$ and velocity $\mathbf{\dot{x}_{\text{des}}}$ is used to determine the $\mathbf{\ddot{x}_{\text{des}}}$ as: 
\setlength{\abovedisplayskip}{4pt}
\begin{equation}
\mathbf{\ddot{x}_{\text{des}} = K_p (x_{\text{des}} - x) + K_d (\dot{x}_{\text{des}} - \dot{x})},
\end{equation}
\setlength{\belowdisplayskip}{4pt}

\subsubsection{Optimization Formulation}
Since Cosmo’s desired task-space objectives exceed its actuated degrees of freedom, a compromise is required. This issue is addressed by solving a weighted optimization that balances task tracking. In this work, an Implicit Hierarchical Whole Body controller (IHWBC) is employed and formulated as a linear constrained Quadratic Program (QP) as shown below:
\setlength{\abovedisplayskip}{4pt}
\begin{equation}
\begin{aligned}
  \min_{\mathbf{\ddot{q}},\,\mathbf{f}}\quad
    & \sum_{i=1}^{N}
      \bigl\|\mathbf{J}_i\mathbf{\ddot{q}} + \mathbf{\dot{J}}_i\mathbf{\dot{q}} - \mathbf{\ddot{x}}_{i,\mathrm{des}}\bigr\|_{\mathbf{W}_i}^2
      + {\mathbf{W_{\ddot{q}}}}\|\mathbf{\ddot{q}}\|^2
      + {\mathbf{W_{f}}}\|\mathbf{f}\|^2, \\
  \text{s.t.}\quad
    & \mathbf{S}_{f}\bigl(\mathbf{H}\mathbf{\ddot{q}} + \mathbf{C} - \mathbf{J}_{c}^\top \mathbf{f}\bigr) = \mathbf{0},\\
    & \mathbf{G}_{f}\,\mathbf{f} \le \mathbf{0},\\
    & \boldsymbol{\tau}_{\min} \le \mathbf{u} \le \boldsymbol{\tau}_{\max}.
\end{aligned}
\end{equation}
\setlength{\belowdisplayskip}{4pt}

The optimization solves for the desired joint accelerations $\mathbf{\ddot{q}}$ and contact forces $\mathbf{f}$. $\mathbf{J}_i$ denotes task space Jacobian matrices. The first terms of the cost function minimizes the weighted task space acceleration errors based on the importance of each task $\mathbf{W}_i$, while the rest two regularization terms on the decision variables encourage minimal control effort. The first constraints enforce rigid-body dynamics to ensure dynamic consistency. A friction cone constraint $\mathbf{G}_f$ bounds the contact forces within feasible limits, while joint torque limits on $\mathbf{u}$ guarantee that the computed solution remains physically realizable on the robot.   

\subsubsection{Inverse Kinematics}
In addition to torque commands, joint position and velocity commands are computed to improve tracking performance. For position control, IK is employed. Due to the complexity of solving analytic IK, a numerical approach using the Damped Least Squares (DLS) method \cite{wampler_manipulator_1986} is adopted. The joint position update is given by
\setlength{\abovedisplayskip}{4pt}
\begin{equation}
    \Delta \theta = \mathbf{\tilde{J}^\top} \mathbf{(\tilde{J}} \mathbf{\tilde{J}^\top} + \lambda^2 \mathbf{I})^{-1} e, \quad \mathbf{\tilde{J}} = \mathbf{WJ},
\end{equation}
\setlength{\belowdisplayskip}{4pt}
where $\mathbf{W}$ is a weighting matrix and $\lambda$ is a damping factor. 

For velocity control, the pseudo-inverse Jacobian method is used, which provides robustness against singularities where the Jacobian may lose rank. The desired joint velocities are calculated as
\setlength{\abovedisplayskip}{4pt}
\begin{equation}
    \boldsymbol{\dot{\theta}}_{\text{des}} = \mathbf{J}^\top (\mathbf{J}\mathbf{J}^\top)^{-1} \boldsymbol{\dot{x}}_{\text{des}},
\end{equation}
\setlength{\belowdisplayskip}{4pt}
\subsubsection{Joint Torque Control}
Given the desired joint position, velocity, and feedforward torque from previous controllers, the final torque command is computed using a PD controller with a feedforward term:
\setlength{\abovedisplayskip}{4pt}
\begin{equation}
\mathbf{u} = \mathbf{K}_p (\boldsymbol{\theta}_{\text{des}} - \boldsymbol{\theta}) + \mathbf{K}_d (\boldsymbol{\dot{\theta}}_{\text{des}} - \boldsymbol{\dot{\theta}}) + \boldsymbol{\tau}_{\text{des}}
\end{equation}
\setlength{\belowdisplayskip}{4pt}

This torque is sent to the motor drivers for execution.

\subsection{Upper Body Motions}
We implemented a Finite State Machine (FSM) for human-like upper body motions, providing a structured approach to managing complex behavioral states and transitions required for realistic humanoid movement and audience interaction.

\subsubsection{Motion Manager}

The FSM governing the robot’s upper body behavior comprises five distinct states, each tailored to different operational needs. The \textbf{REST} state serves as a calibrated reference posture, ensuring the robot consistently returns to a known configuration. In certain modes, it can also initiate idle behaviors that mimic a fidgeting child waiting, adding a layer of relatability and realism. The \textbf{ACTION} state is responsible for executing predefined motion sequences, encompassing the robot's primary gestures and upper body movements. The \textbf{TRANSITION} state interpolates between motion states. For moments requiring stillness—such as photo opportunities—the \textbf{STOP} state allows the robot to pause mid-motion and resume precisely from where it left off. Finally, the \textbf{JOYSTICK} state grants manual control of the head joints, facilitating interactive experiences with the audience or enabling tracking of specific visual targets.

\subsubsection{Motion Planning}

Smooth transitions between FSM states are achieved using cubic Hermite splines, which interpolate between the current (position, velocity) and target states during the TRANSITION state. This method ensures  C¹ continuity, maintaining smooth and continuous velocities throughout the trajectory.

When implementing this approach, our system first enters the TRANSITION state upon trigger, captures current joint positions and velocities as initial conditions, and identifies target values from the destination state. The cubic Hermite spline parameters are then calculated and executed until the interpolation is complete, at which point the FSM transitions to the target state. This ensures all state changes occur through smooth, controlled trajectories rather than abrupt movements.




%% file: Sections/S4_results.tex
\section{Results} 
\label{sec:experiment results}
Cosmo was evaluated through a series of hardware and software experiments to highlight its design features and demonstrate its capability for dynamic locomotion combined with expressive upper body motions.

\begin{figure}[t!]
    \centering
    \includegraphics[width=1\linewidth]{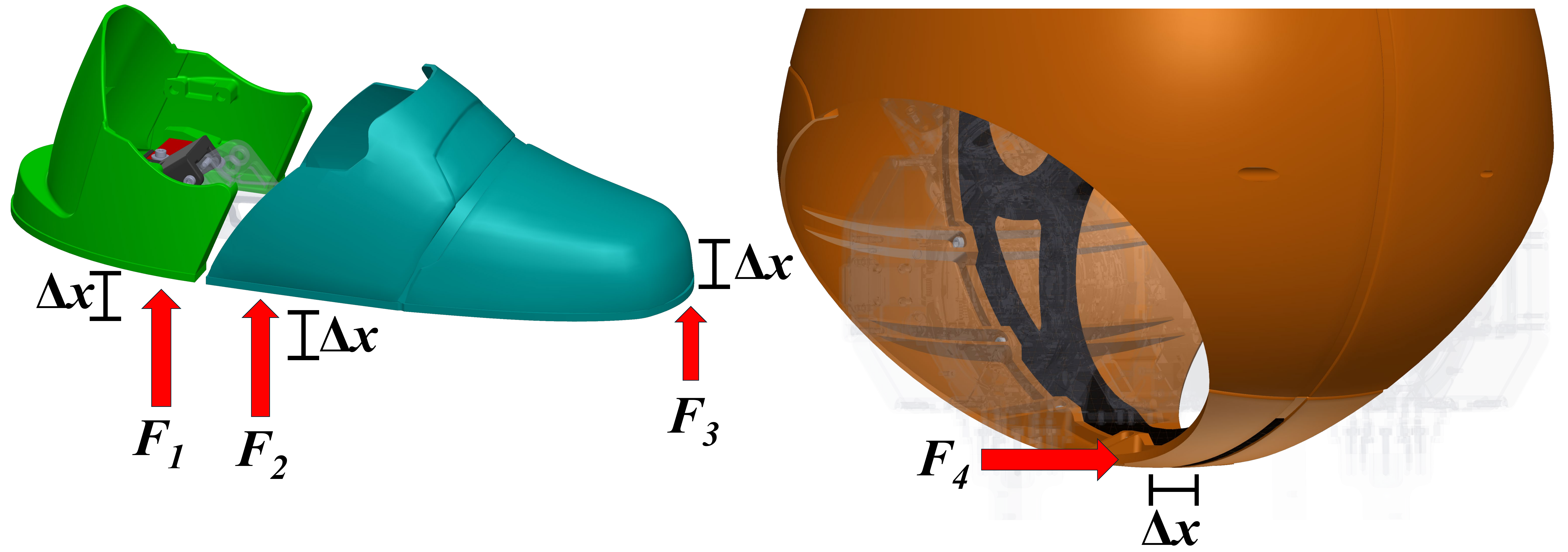}
    \caption{Locations and directions of applied forces during compliance testing. Boot shells were displaced $\Delta$x = 4, 8, 12, and 16 mm in each axis. The body shells were displaced $\Delta$x = 1.75 mm and 3.5 mm.}
    \label{fig:compliance_figure}
\end{figure}

\begin{table}[t!]
 \centering
 \caption{Shell Stiffness Results}
 \begin{tabular}{ c c c c c }
    & \textbf{Heel} & \textbf{Toe Roll} & \textbf{Toe Pitch} & \textbf{Body} \\
    & \textbf{(F1)} & \textbf{(F2)} & \textbf{(F3)} & \textbf{(F4)} \\
    \specialrule{.15em}{.05em}{0em}
    $k_x$ [N/mm] & 0.123 & 0.147 & 0.098 & 5.774 \\
    
 \label{table:compliance_table}
 \end{tabular}
\end{table}

\subsection{Shell Compliance}
To validate our boot shell compliance and body shell impact mitigation, the forces required to move the shells a set of displacements were recorded, from which stiffness values are calculated. All boot axes of compliance in the boots were tested to 16 mm, a displacement further than operationally expected. Even at this displacement, our recorded force never exceeded 2.4 N or 1\% of the robot's weight. Thus, we are confident boot displacement will not generate forces significant enough to impact our locomotion. Body shell compliance is noticeably stiffer compared to the boots and allows for less displacement. However, the compliance is sufficient to absorb rare impact from the femur during locomotion. The location and direction of applied forces can be seen in \cref{fig:compliance_figure} while the calculated stiffnesses are shown in \cref{table:compliance_table}.

\begin{figure}[t!]
\centering
\includegraphics[width=0.99\linewidth]{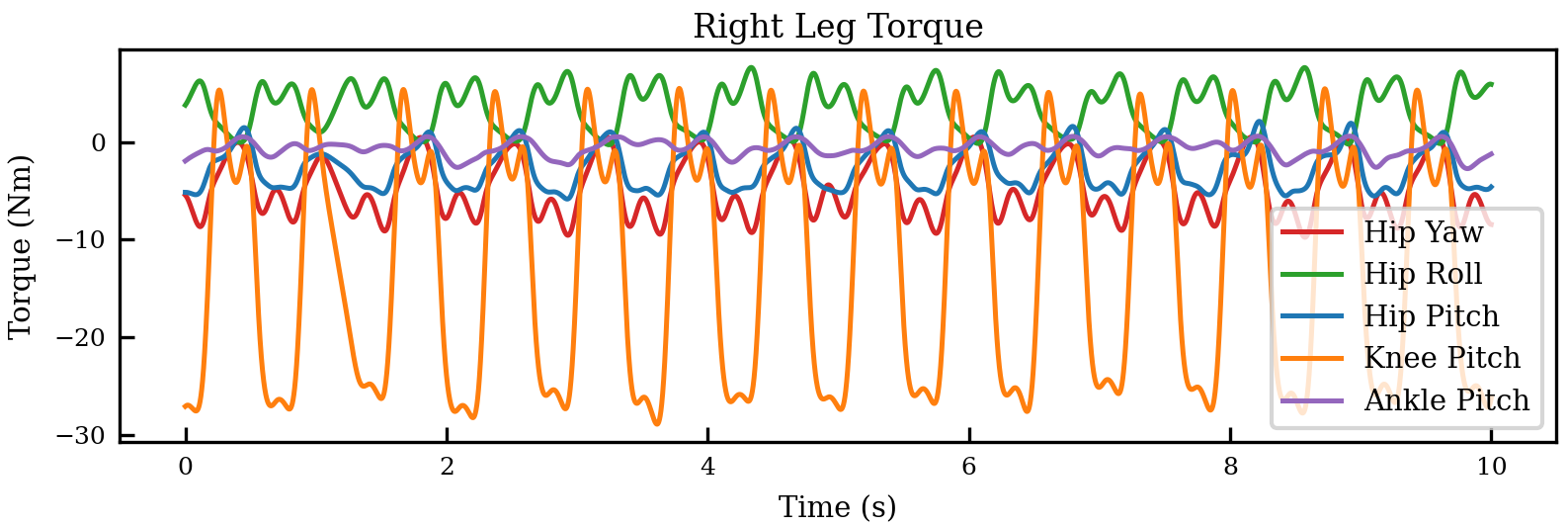}
\caption{Measured torque profiles of the right leg joints during stepping.}
\label{fig:motor_torque}
\end{figure}

\subsection{Joint Torque Evaluation}
To validate the hardware design and joint-level control performance, motor torque profiles were recorded during stepping. As shown in Fig.~\ref{fig:motor_torque}, the 45° and 135° hip joint configuration effectively distributes loads between the yaw and roll axes. The average torque magnitudes were 3.93 Nm for hip yaw and 3.98 Nm for hip roll, confirming balanced load sharing. Additionally, the torque data show smooth transitions and low-frequency oscillations throughout the gait cycle, indicating minimal backlash or compliance in the transmission.

\begin{figure}[htbp]
\centering
\includegraphics[width=0.99\linewidth]{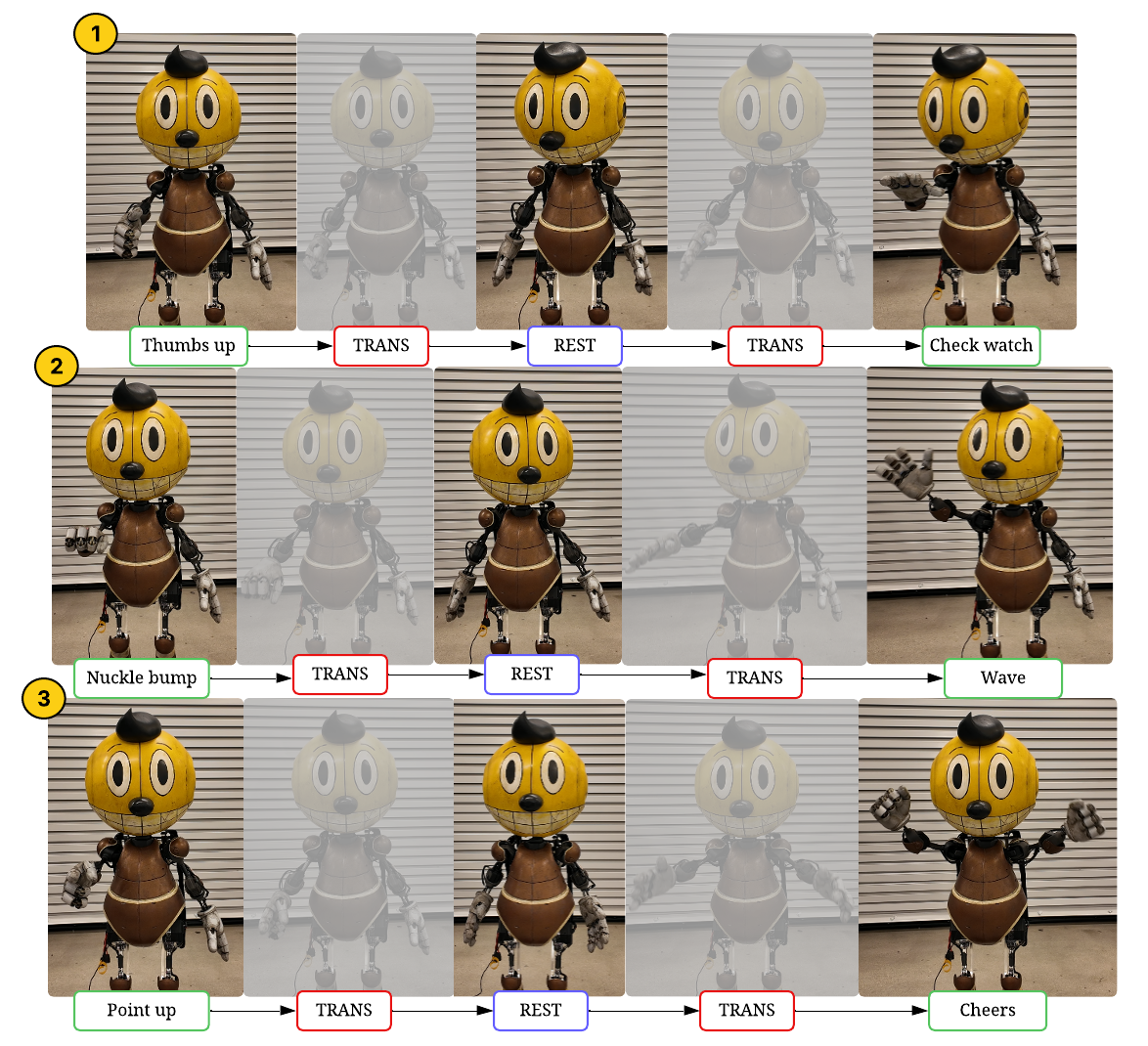}
\caption{From top to bottom: FSM states illustrating the system's state transitions across three sequences while standing. Green blocks represent ACTION states, red blocks show TRANSITION states, and purple indicates REST states. The sequential transitions between these states demonstrate the robot's ability to perform discrete gestures while maintaining stable balance.}
\label{fig:up_results}
\end{figure}

\subsection{Upper body motions}
We evaluated upper body gesture control during stationary standing, using a constant transition time (t = 1) between states. Figure \ref{fig:up_results} shows our FSM-based motion system in action, with green blocks indicating ACTION states, red blocks showing TRANSITION periods, and purple representing REST states. Our recorded action library enables effective human-robot interaction suitable for public demonstrations. The state machine smoothly coordinates transitions between different poses (thumbs-up, rest positions, cheering), creating natural motion sequences while protecting hardware from damage during rapid changes. 

\subsection{Locomotion Framework Implementation}
The WBC employs the commercial available QP solver PROXQP \cite{bambade_proxqp_2023}, with the complete QP formulation and control loop running at 1 kHz. Real-time performance is ensured by synchronizing the state estimator and hardware interface at the same rate, while user commands and reference trajectory planners operate at a lower frequency, between 100 Hz and 500 Hz.

\begin{figure}[htbp]
\centering
\includegraphics[width=0.95\linewidth]{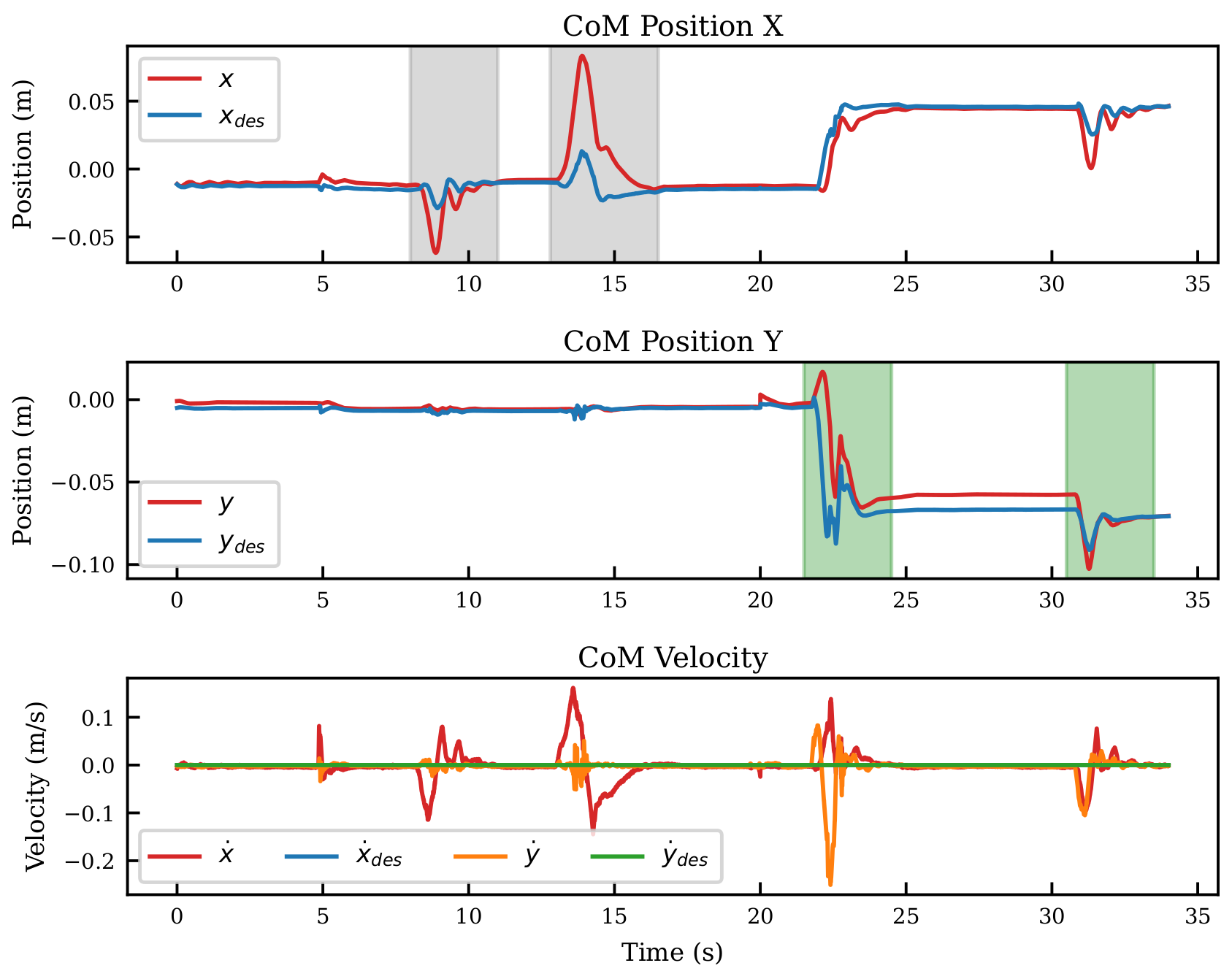}
\caption{CoM tracking during push recovery. The top and middle plots show CoM position in the X and Y directions; the bottom plot shows velocity. Grey and green shaded regions indicate pushes in the X and Y directions, respectively.}
\label{fig:balancing}
\end{figure}

\subsection{Disturbance Rejection During Balancing}
To assess Cosmo’s balance robustness, we conducted a disturbance rejection test with the robot standing under constant reference inputs. Manual perturbations were applied to the torso in both the sagittal ($X$) and lateral ($Y$) directions. As shown in Fig.~\ref{fig:balancing}, the CoM displaced up to 8 cm in both directions but returned to a stable posture without falling. Velocity plots show a settling time under 2 seconds, defined by speeds below 0.01 m/s. The desired CoM position is set at the midpoint between the feet. In response to disturbances—visible in the initial $Y$-direction push—the robot steps to expand its support polygon and shifts the CoM reference to maintain balance.


\begin{figure}[t!]
\centering
\includegraphics[width=0.95\linewidth]{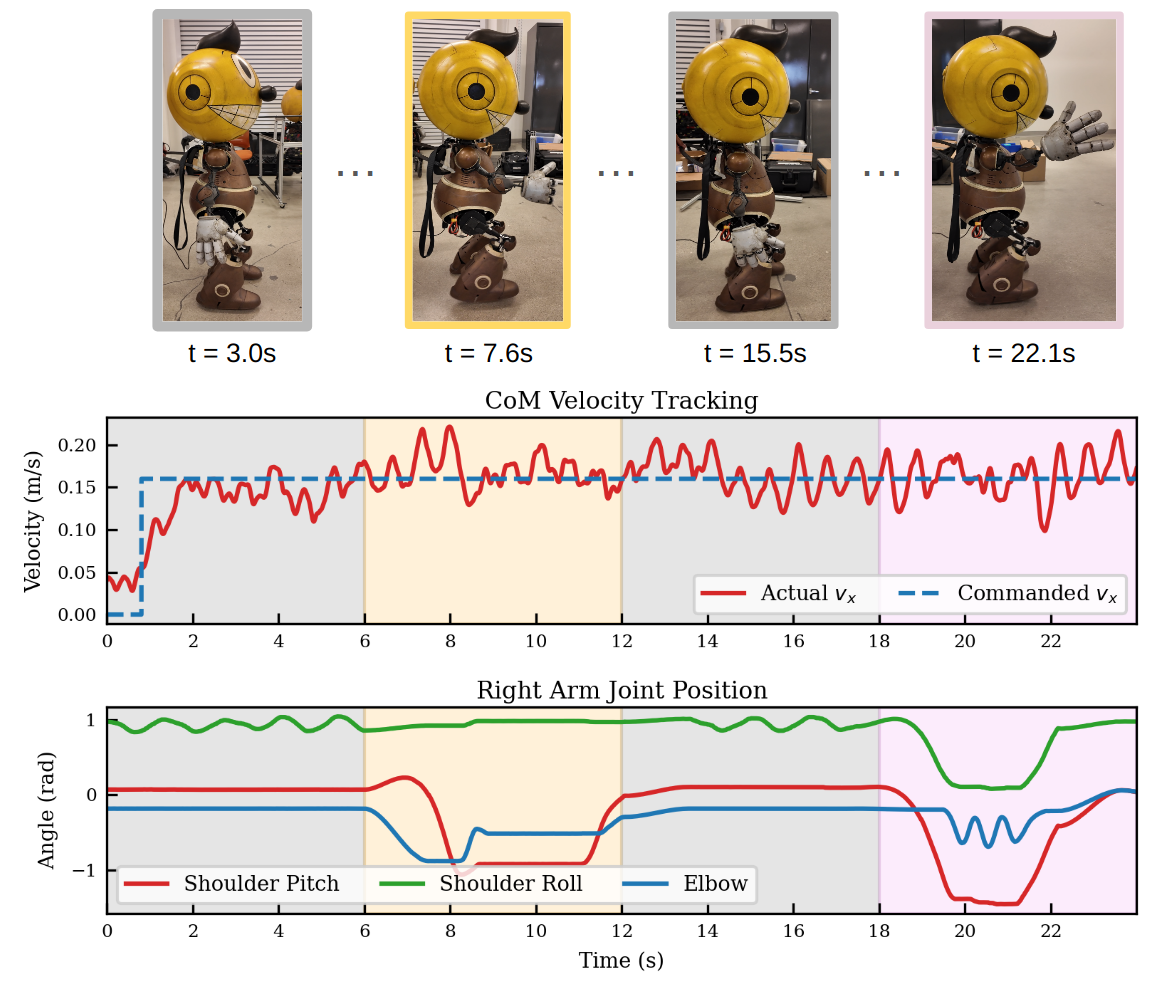}
\caption{Experimental results of Cosmo's walking while performing upper body gestures: idle (gray), thumbs up (orange), and waving (pink). The top row shows snapshots of each phase. The plots below show forward velocity tracking and right arm joint trajectories reflecting gesture patterns}
\label{fig:walking_w_upperbody}
\end{figure}

\subsection{Walking With Upper Body Movements}
To evaluate Cosmo’s ability to maintain stable locomotion while performing upper body gestures, the robot walked at a constant 0.16 m/s while executing a sequence of motions. The right arm joint trajectories exhibit distinct gesture patterns, highlighted in Fig.~\ref{fig:walking_w_upperbody} with shaded regions and snapshots. Throughout the experiment, the forward CoM velocity closely tracked the command, with an average deviation of 0.012 m/s. These results demonstrate that Cosmo’s whole-body torque control enables stable walking with large, time-varying upper body movements, with minimal impact on velocity or stability.



%% file: Sections/S5_conclusion.tex
\section{Conclusion} 
\label{sec:conclusion}
This work presented Cosmo, a humanoid robot designed to replicate a fictional character while achieving robust, torque-controlled locomotion and expressive upper body motions. Experimental results demonstrated Cosmo’s ability to walk stably while executing various gestures and to reject external disturbances during standing. These findings highlight the potential of performance-oriented humanoid robots that integrate character embodiment with real-world mobility. Future work will explore more dynamic behaviors, interactive capabilities, and expressive learning-based methods.